# Custom Video-Oculography Device and Its Application to Fourth Purkinje Image Detection during Saccades

Evgeniy Abdulin[†]
Computer Science Department
Texas State University
San Marcos, TX, USA
abdulin@txstate.edu

Lee Friedman
Computer Science Department
Texas State University
San Marcos, TX, USA
lfriedman10@gmail.com

Oleg Komogortsev
Computer Science Department
Texas State University
San Marcos, TX, USA
ok@txstate.edu

## ABSTRACT

We built a custom video-based eye-tracker that saves every video frame as a full resolution image (MJPEG). Images can be processed offline for the detection of ocular features, including the pupil and corneal reflection (First Purkinje Image, P1) position. A comparison of multiple algorithms for detection of pupil and corneal reflection can be performed. The system provides for highly flexible stimulus creation, with mixing of graphic, image, and video stimuli. We can change cameras and infrared illuminators depending on the image qualities and frame rate desired. Using this system, we have detected the position of the Fourth Purkinje image (P4) in the frames. We show that when we estimate gaze by calculating P1–P4, signal compares well with gaze estimated with a DPI eye-tracker (which natively detects and tracks the P1 and P4).

## 1 INTRODUCTION

Video-oculography or VOG, also known as video-based eye tracking, is the dominant method today for measuring eye-position. Although it is quite accurate, further increases in accuracy may be possible. Some have claimed that it cannot be used to accurately measure saccade characteristics accurately (Hooge, Holmqvist, & Nyström, 2016). VOG is also susceptible to various artefacts, such as noise (E. Abdulin, Friedman, & Komogortsev, 2017). We have developed a custom VOG system including hardware and software components, with new characteristics that will enhance the search for methods to improve the accuracy and robustness of eye position estimation.

One approach to improving VOG is to improve the detection of the positions of the pupil and corneal reflection, which are central to the traditional estimate of eye position. The ability to save high-resolution video frames at high temporal precision for offline analysis provides the data necessary to explore new detection methods. Another approach to improvement of the VOG estimate of eye position is to include also the detection of other ocular and periocular features. The third approach is to develop alternative methods of measuring eye position, by modeling, for example, a Dual Purkinje Image (DPI) eye tracker. For this, the Fourth Purkinje image (P4) would need to be detected, something we demonstrate with our device below.

Any new eye tracking device must be evaluated in terms of tradition performance metrics, such as accuracy, precision, linearity, crosstalk and sample rate (temporal) stability.

In the present report, we describe the characteristics of our device, including the many new capabilities it provides. Its performance characteristics is fully analyzed. Its potential by detecting the Fourth Purkinje Image, and creating a modeled DPI signal, is demonstrated. We also demonstrate the ability to perform multiple eye position estimations of from the same video stream with the use of varying pupil-intensity detections threshold.

## 2 PRIOR WORK

Despite of the domination of VOG systems, using of DPI systems and P4 detection is still employed in recent studies, such as (Han, Saunders, Woods, & Luo, 2013; Zhang et al., 2008). (Nyström, Hansen, Andersson, & Hooge, 2016) also employs P4 for microsaccades study, our study includes P4 detection and tracking during regular saccades. Some of the published papers are worth extra attention, such as (He, Donnelly, Stevenson, & Glasser, 2010), where authors employs a DPI eye tracker but also needed to use a video camera as an extra device to estimate pupil diameter. In (Bueno, De Brouwere, Ginis, Sgouros, & Artal, 2007) P4 is analyzed by a video camera as well. Since our eye tracker is designed to capture video frames from the camera, the listed tasks can be fulfilled without any modifications.

## 3 CUSTOM EYE TRACKER

We built a non-commercial eye-tracker with the main purpose to get several key capabilities, which are mostly unavailable in commercial devices. These capabilities are provided by designing of the eye tracker as an integrated system, where all of its components are carefully chosen based on compatibility and its role.

- Saving every frame at its full resolution at framerates up to 500fps. This allows for off-line analysis of various ocular and periocular features and the evaluation of various signal types, e.g. gaze estimation or motion of the features. Specifically, in the present case, we created a signal representing the position of the P4 from VOG data. The characteristics of frame processing algorithm provides high time stability of sampling rate. The frames are recorded with the intra-frame MJPEG format, rather than in inter-frame formats, which save completely only selected frames (base frames) and the most of the frames are saved as differences of subsequent frames from a particular base frame.
- Tailoring the pixel dimensions of the video frame to each application (e.g., monocular, binocular recordings, extended view of periocular features and others). This also provides trading-off between video frame resolution and frame rate with high precision – up to 1 screen pixel of frame resolution and 1 fps of frame rate. Thus, those parameters can be set specifically to meet the goals of a particular experiment.
- Programmable stimulus. We developed a XML-based set of commands that can be used to program the stimulus. The stimulus can be presented as dot on a homogeneous background, images (pictures, text pages and other) or video streams. For dot stimuli, our system supports enhanced capabilities: specifying target position directly in degrees of visual angle rather than the more common pixel units, random fixation times, various velocity in transitions, shrinking in fixations.
- Swappable optical components: IR light sources with various geometry, size, power and wavelength; cameras with various characteristics of the sensor and control circuits to control pixel resolution, frame rate, sensitivity to IR light; optics for the cameras – lenses and filters.
- One of our IR illuminators is designed in our laboratory and includes: (a) three modes of light intensity controlled by a switch; (b) glitch-free light stream that is provided by battery power; (c) 940nm IR light that is invisible for human eye when turn on preventing distractions during recordings; (d) omnidirectional stand.

For the data described below, our system consisted of the following components, see Fig. 1.

(1) Our modification of an open-source eye-tracking algorithm ITU GazeTracker (Skovsgaard, Agustin, Johansen, Hansen, & Tall, 2011) to process video frames at full resolution without frame losses. (Our modifications of the ITU GazeTracker code is explained in Appendix) The eye tracker camera is connected to the desktop through USB 3.0 and the desktop is not shown in Fig.1

(2) Camera IDS-3180CP2-M with lens (Kowa LM35SC lens with MidOpt LP780 visible light filter). If needed, the camera IDS-3370CP2-NIR with lens Kowa LM35HC and MidOpt LP800 filter can be installed. All of the optical components – IR light sources, mirror, filters, lenses and cameras' sensors are selected to provide high performance in 850nm and 940nm infrared light.

(3) IR light source with 850nm wavelength. Can be replaced by another IR light source with 940nm wavelength with.

(4) Hot mirror glass, which is transparent for visual light and reflective for infrared light. The mirror is mounted on an adjustable stand, manufactured in special engineering facility with our drafts. The hot mirror which is placed between the display and the participant's eyes and provides a capability to position the camera virtually right in front of the eyes.

(5) Computer monitor to display a visual stimulus. Its screen dimensions are 374x300 mm and resolution is 1280x1024 pixels. The monitor is placed at the distance of 500mm from the participants' eyes. Cyclopean axis ia aligned to monitor's center.

(6) Dell Alienware X51 R3 desktop with Intel Core i7-6700K CPU at 4GHz and 16GB RAM. The CPU is selected to be able to process the video stream at frame rate provided by a high-speed cameras.

(7) Chin and front rest to stabilize a participant's head.

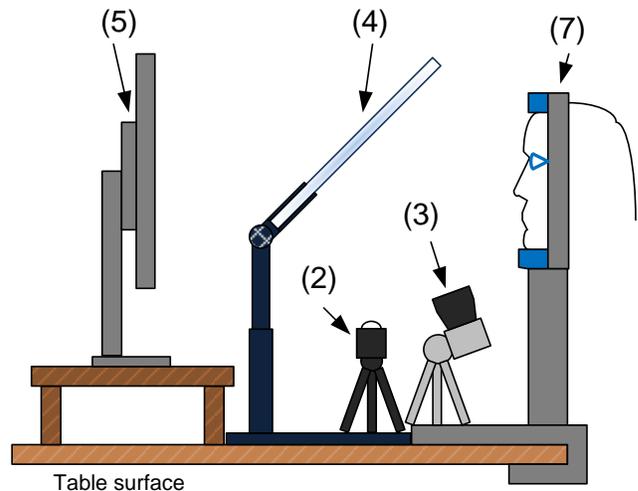

Figure 1: Placing of the eye tracking setup key components

Our setup has two modes – record and replay. In the record mode, before starting an actual recording, the ITU GazeTracker reads and processes the program stimulus file and calculates all of the timings including the ones that are programmed to be random. Detailed target information is written into an XML file. During recording, the camera captures frames and places them in a software buffer. The main priority of the ITU Gaze Tracker in this mode is to record frames to a MJPEG video file without dropping any frames. As a secondary task, ITU GazeTracker computes a temporary estimate of gaze position that is monitored by an operator to ensure correct operation of the applied eye-tracking algorithm and resulting data quality. When the recording is finished, video frames from the recording, including the calibration period, and calibration model data, are stored in corresponding files on the desktop.

In the replay mode, the video frames are read from pre-recorded video files. The primary task of this mode is to find the Pupil and the P1 in each frame and calculate their positions, create a difference signal (Pupil–P1) and calibrate it to produce gaze position samples. The samples are recorded along with other diagnostics information. Since this analysis is done off-line, it is possible to perform any analysis that is necessary since there is no time limitations for that. For example, we could compare several different Pupil or P1 detection algorithms or different calibration methods on exactly the same video stream of the eye movement. This provides better experimental purity of the comparison results in contrast with comparison of different processing of different records.

## 4 METHODS

Our study has two goals: (1) Characterization of performance of our device and (2) Demonstration of its enhanced research capabilities.

*Participants* We used records from 5 male participants, mean ages 21-34 (M = 24.2 years, SD = 5.5 years), members of our research team who provided an informed consent for the recordings. For the first task, the data from only the first participant, a 23-years old student, was used. For the second task, we used the data from all 5 participants.

*Characterization of Performance.* The linearity of an eye-tracker can be assessed by relating final measures of eye-position to target position. However, when measured at this level, the linearity could be strongly influenced by the transformation by implemented calibration model, which may include quadratic and higher order components. Another, perhaps more meaningful, linearity test would relate the signal prior to calibration to target position. In the case of VOG, this would involve comparing the corresponding Pupil–P1 signal component to target position components. In this case, crosstalk check would use comparison the same signals, but, in contrast to the linearity check, it will use orthogonal components, horizontal vs vertical and vice versa. The majority of commercial eye trackers do not provide data for linearity and crosstalk checks. Our device provides all the data necessary to perform these evaluations. We used a resolution of 512x320 pixels with a sampling rate of 500 Hz. Note that this is a higher spatial and temporal resolution than that employed by (Nyström et al., 2016). In their prior study, they used a resolution of 320x200 and a sampling rate of 240 Hz.

Accuracy measurements were performed in the record where target position was determined, as a Euclidean distance between centroids of recorded gaze positions during fixations and corresponding target positions. Spatial precision measurements were done based on intersample distances (Holmqvist, Nyström, & Mulvey, 2012).

The visual stimulus was presented as a single black dot on light gray background. The dot's parameters were d=0.67 ° with jumping movements either to the left or right, forming pure horizontal saccade targets, or up and down, forming pure vertical saccade targets. Dwell time was random from 1 to 2 sec. for all dot positions. We tested a range of amplitudes from 2.5 to 40° for horizontal saccades and from 2.5 to 30° for vertical saccades with step of 2.5°.

*Demonstration of Enhanced Capabilities.* To illustrate the enhanced capabilities of our system we develop methods to detect P4 in our VOG system. The details of this detection procedure are described below. Thus, our study includes joint analysis of activity of three features during saccade eye movements – pupil center, P1 and P4 to be able to obtain both signals, from VOG and modeled DPI eye tracker. We compare examples of eye-position from a true DPI tracker (Crane & Steele, 1985) from the prior study (Deubel & Bridgeman, 1995) with our model of a DPI signal using P1 and P4 positions estimated by our system, as a simple difference vector P1–P4, whereas VOG system uses Pupil–P1 difference vector. For the consistency purposes, the task for the study replicated the task for the previous study of comparison DPI eye tracker with scleral search coils and contains several sequential saccade stimulus for pure horizontal saccades at 4° and pure vertical saccades at 15° of visual angle.

To demonstrate the ability to apply multiple analyses to the same frame, we show how changes in the pupil detection threshold affect gaze estimates.

In addition, to verify the ability of accurate and reliable detection and tracking of Pupil, P1 and especially P4, we recorded 10° horizontal saccades with multiple participants.

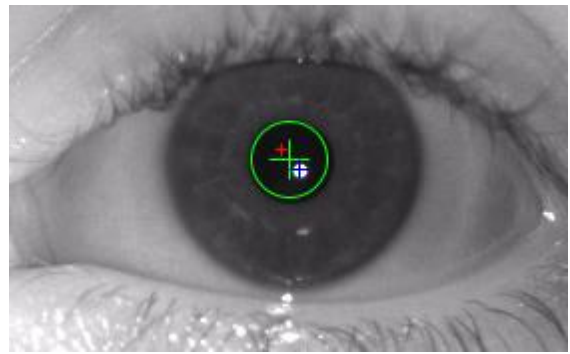

**Figure 2: Example of features detection on the frame for the camera: Pupil (green), P1(blue), P4(red).**

There were a number of challenges in the detection of P4 (Fig. 3).

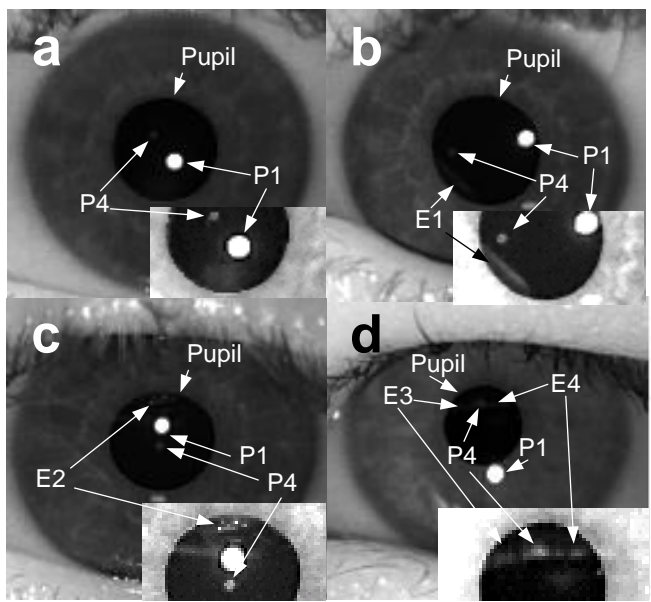

**Figure 3:** Various issues in P4 detection, miniatures in right bottom corners have amplified brightness for better visibility, a – no issues, P4 is easily detectable; b – E1 – extra reflection from participant's nose; c – E2 – occlusions and reflections from participant's eyelashes; d – the most difficult case, P4 is among extra reflections E3 and E4 from supraorbital area.

## 5  ALGORITHM OF FOURTH PURKINJE IMAGE DETECTION AND TRACKING

**Input:** A video frame from a recorded video stream and a Pupil position and size, estimated by VOG algorithm for that frame, to find a limit Area of Interest (AOI) for search of P4; we also used estimation of P1 position, to exclude P1 from P4 search.

**Step 1:** Forming AOI by removing all pixels beyond the Pupil area, using estimated Pupil position and size, from consideration using masking process.

**Step 2:** Pre-processing of AOI by removing too faint pixels using an adaptive threshold and too bright pixels using fixed threshold (intensity>65, intensity range of all images is from 0 to 255). The adaptive threshold is estimated based on pixel intensity values of AOI, taking into account noise and intensity of extra reflections in the Pupil area, shown in Fig.3. The threshold depends on individual's eye properties.

**Step 3:** Use the `regionprops` algorithm in MATLAB, to automatically detect relatively high intensity blobs within each pupil AOI, using 8-connectivity rule. For each blob, the algorithm omputes a centroid position, area, mean intensity and other parameters.

**Step 4:** The transition path of P4 for regular saccades traverses various areas of pupil with inhomogeneous brightness due to extra reflections from periocular areas and individual pupil features. The task of P4 detection is more complex in these conditions and requires more advanced algorithms (Hansen & Ji, 2010), such as a Feature-based method. Thus, to be considered a potential P4, a blob must have an area between 5 and 30 pixels and a maximum brightness >= adaptive threshold, which depends on individual's eye properties and is computed based on measured intensity values of detected blobs and another adaptive threshold value from Step 2. From all of the blobs which meet these criteria, we chose the blob that is closest one, by calculating its Euclidean distance to the pupil center.

## 6  RESULTS

*Characterization of Performance.* Our linearity results are presented in Fig. 4. Signal noise and loss can result during blinks and RIONEPS artifact (E. Abdulin et al., 2017). For this particular recording, from this subject, after removing the noise, the data validity was 90%.

The VOG data quality of the record can be described by following parameters:

- Measured accuracy after calibration, average error: 0.50 °.
- Measured accuracy in recording across 190 fixations in different places of the monitor's screen: M = 0.54°, STD = 0.38°.
- Measured spatial precision for the same fixations: M = 0.12°, STD = 0.01°.

Samples were either 2.0030 or 2.0031 msec apart. The proportion of intersample intervals that were 2.030 msec apart is 0.47, the proportion of intersample intervals that were 2.0031 msec apart is 0.53. The average intersample interval was 2.0030398548 which is a factual sampling rate of 499.2412 for set sampling rate of 500Hz. There was a very strong tendency for a 2.0030 to be followed by a 2.031 and vice versa. This was demonstrated by a highly statistically significant lag 1 temporal autocorrelation of −0.9 (p =0.0). Note that although this autocorrelation is very high, it is not 1.0.

The linearity check of our device shown that the device is linear across the entire range for horizontal and vertical signal,

Results of crosstalk check are presented on Fig. 5. Based on calculated slope values, for horizontal signal, the crosstalk is less that 1%, and for vertical signal it is less than 4% since the detection of vertical eye movement is more difficult.

*Demonstration of Enhanced Capabilities.* Out P4 detection results are illustrated in Fig. 2,

As it shown on Fig. 7, our modeled DPI position signals (shown in blue, (c) and (d)) are consistent in general with the DPI signals of (Deubel & Bridgeman, 1995) (Figure 7, (a) and (b)). Both the true and our modeled DPI signals have similar saccadic overshoots, which do not appear in our traditional VOG signal, shown in red. For our modeled DPI signals, there is a dip just prior to saccade onset which is not present in the true DPI signal. We have no ready explanation for this difference at this time. Perhaps this is subject specific, or related to some other detail in the recording method. The data from Deubel and Bridgeman (1995) was recorded while a scleral search coil was in place on the eye.

Custom Video-Oculography Device and Its Application to Fourth Purkinje Image Detection during Saccades

The capability of off-line analysis of the exactly same record under various conditions, such as different threshold values, is shown on Fig.6. Increasing and decreasing the values of Pupil Detection Threshold affects the level of noise. For example, if after the recording the employed threshold seems to be poorly estimated, this can be fixed by simple re-processing the video with better threshold value, it is especially important for signals affected by noise.

Recorded saccades estimated by VOG and modeled DPI signals shown in Fig. 8., indicate the capability of our eye tracker to detect and track Pupil, P1 and P4.

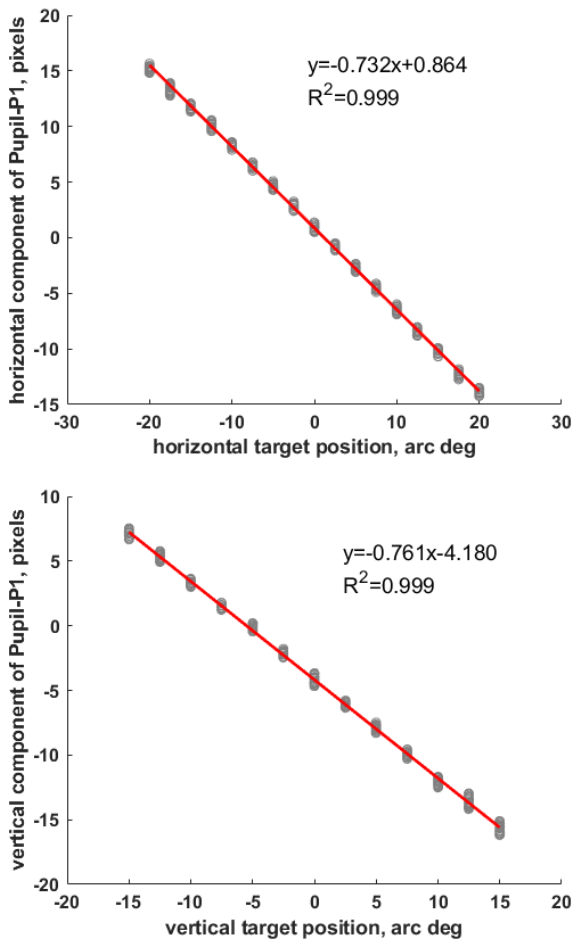

Figure 4: Linearity of our eye tracker for horizontal and vertical gaze position signals

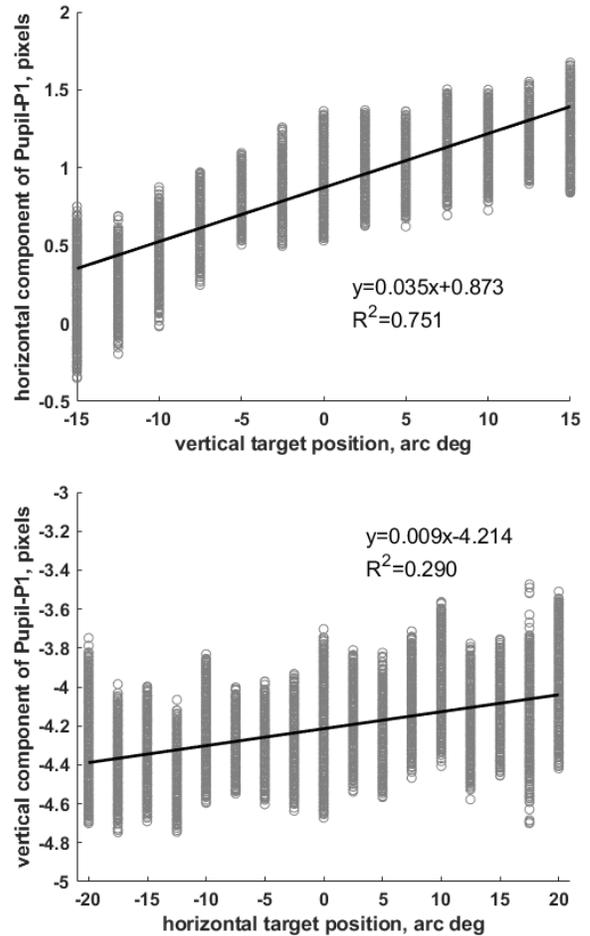

Figure 5. Crosstalk of our eye tracker for horizontal and vertical gaze position signals

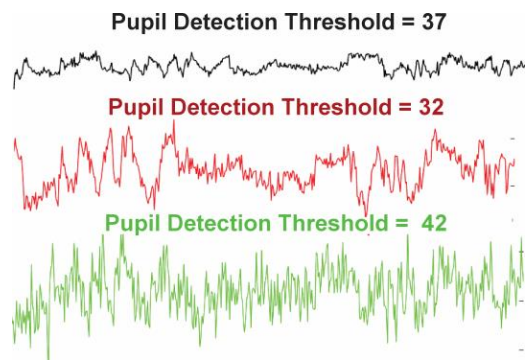

Figure 6. Estimated gaze position signal depending on Pupil Detection Threshold, best fit = 37, increased = 42, decreased = 32 from the same record

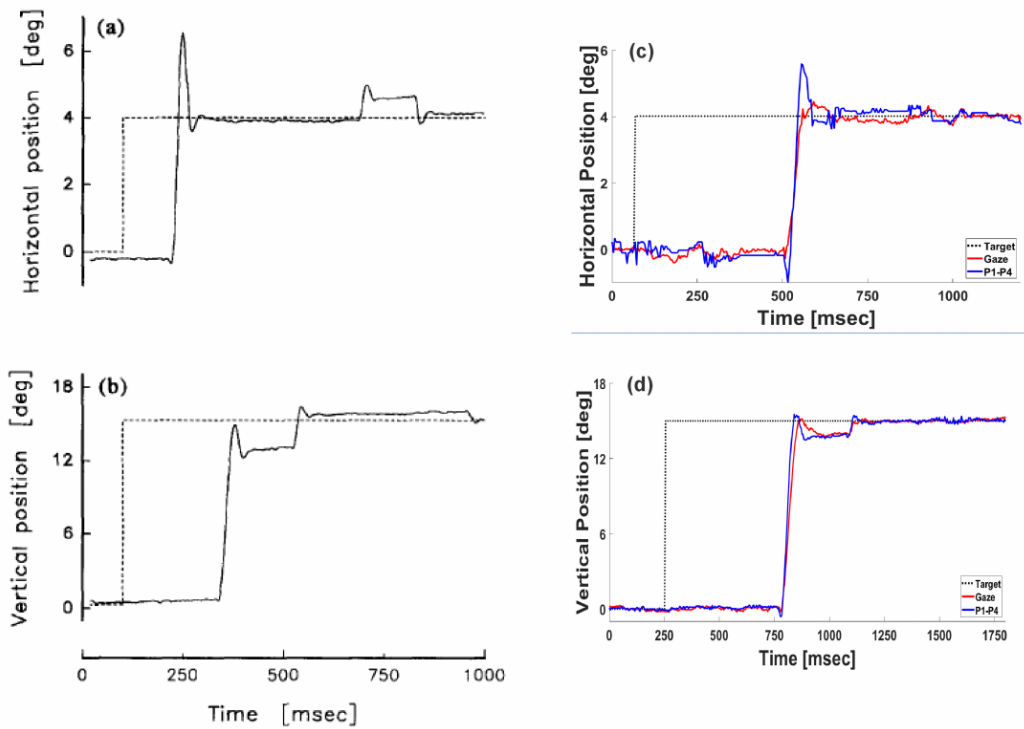

**Figure 7: Comparison of saccade detection by DPI eye tracker (on the left) (Deubel & Bridgeman, 1995) and new eye tracking setup (on the right)**

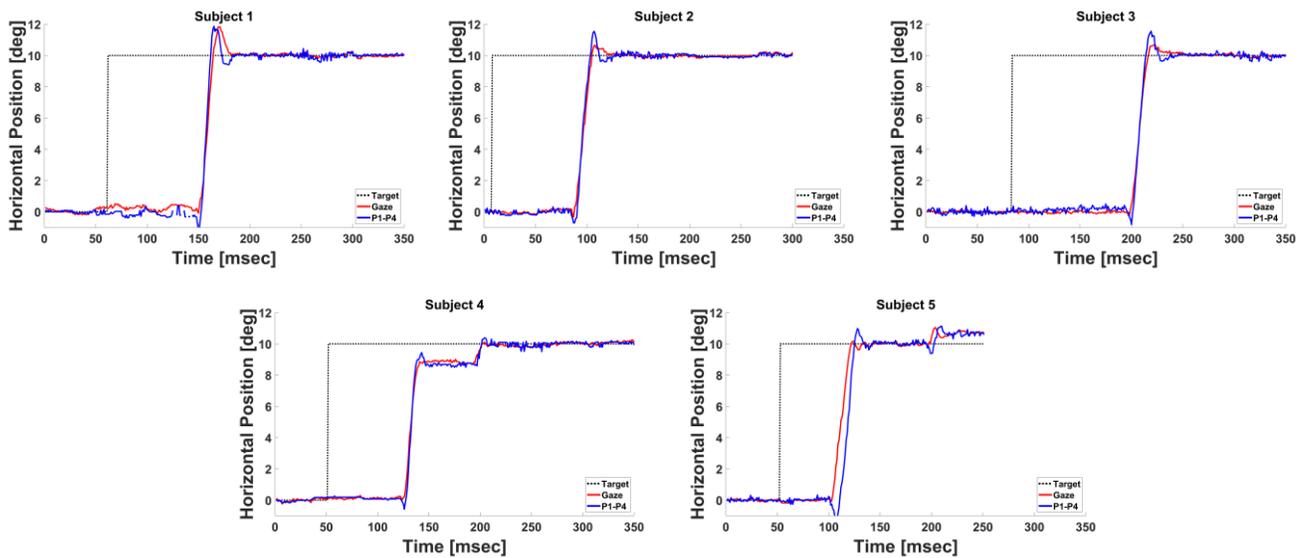

**Figure 8: Leftward 10° saccades estimated by VOG and modeled DPI signals based onetected Pupil, P1 and P4, from multiple participants**

## 7  DISCUSSION

In this report, we described a custom-built VOG eye-tracking device along with its main features, new capabilities and advantages. These enhancements were illustrated by detecting a new target for VOG, i.e., the Fourth Purkinje Image (P4). The capability of offline reanalysis was demonstrated by displaying the results of the use of different pupil-intensity detection thresholds. Device's accuracy, precision, linearity and crosstalk were evaluated as well.

The ability for offline repeat analyses with high allowance of processing times without limits is the most important capability provided by our device. The exploration of various new approaches to the estimation of eye-position is now possible. We would also point to our very high temporal stability. At present, our system allows for offline analysis only. In the future, we would like to provide for real-time estimation of eye-position.

Our system demonstrated very good linearity and low crosstalk. Accuracy and precision are reasonable, but there is definitely room for future improvement. Enhanced pupil detection algorithms may be the key for this.

In addition to Pupil and P1, we were able to detect accurately and reliably the position of P4 in pre-recorded video frames. In comparison with the prior study (Nyström et al., 2016) also presented the detection of P4 in video frames, but the focus of that paper was a study of microsaccades. Detection of P4 more generally, as we have done, is more difficult because, with larger saccades, the pupil image is more susceptible to other confounding areas of increased intensity due to various reflections and occlusions. In the future, we will try to make our P4 detection more robust, either by improving the logic of detection or applying machine learning methods. Our future work will include planning and conducting the next study with larger number of participants as well.

The detection of P4 is the next step of ocular and periocular feature analysis, previously reported in (E. R. Abdulin & Komogortsev, 2017).

If we can produce a robust and accurate estimate of P4 position, our main experimental emphasis going forward will be to try combine traditional VOG eye-position signals and modeled DPI eye position signals to produce a better estimate of eye position. As we expect, such a signal would be more accurate, precise, and linear and less susceptible to crosstalk and various artefacts.

## 8  CONCLUSION

We have developed a custom eye-tracking device, which opens up new possibilities for further eye-movement measurement research. At this early stage, we have already shown our ability to detect P4 in video-frames. Methods for the improvement of traditional VOG can be explored. Also, we can also evaluate the usefulness of tracking other ocular and periocular features.

## ACKNOWLEDGMENTS

This work is supported by the National Science Foundation under grants #CNS-1250718 and #CNS-1714623. The work is inspired by the Google Virtual Reality Research Award and Google Global Faculty Research Award bestowed on Dr. Komogortsev in 2017 and 2019, respectively.

## REFERENCES

Abdulin, E., Friedman, L., & Komogortsev, O. V. (2017). Method to detect eye position noise from video-oculography when detection of pupil or corneal reflection position fails. *arXiv preprint arXiv:1709.02700.*

Abdulin, E. R., & Komogortsev, O. V. (2017). *Study of Additional Eye-Related Features for Future Eye-Tracking Techniques.* Paper presented at the Proceedings of the 2017 CHI Conference Extended Abstracts on Human Factors in Computing Systems.

Bueno, J. M., De Brouwere, D., Ginis, H., Sgouros, I., & Artal, P. (2007). Purkinje imaging system to measure anterior segment scattering in the human eye. *Optics letters, 32*(23), 3447-3449.

Crane, H. D., & Steele, C. M. (1985). Generation-V dual-Purkinje-image eyetracker. *Applied Optics, 24*(4), 527-537.

Deubel, H., & Bridgeman, B. (1995). Fourth Purkinje image signals reveal eye-lens deviations and retinal image distortions during saccades. *Vision research, 35*(4), 529-538.

Han, P., Saunders, D. R., Woods, R. L., & Luo, G. (2013). Trajectory prediction of saccadic eye movements using a compressed exponential model. *Journal of vision, 13*(8), 27-27.

Hansen, D. W., & Ji, Q. (2010). In the eye of the beholder: A survey of models for eyes and gaze. *IEEE transactions on pattern analysis and machine intelligence, 32*(3), 478-500.

He, L., Donnelly, W. J., Stevenson, S. B., & Glasser, A. (2010). Saccadic lens instability increases with accommodative stimulus in presbyopes. *Journal of vision, 10*(4), 14-14.

Holmqvist, K., Nyström, M., & Mulvey, F. (2012). *Eye tracker data quality: what it is and how to measure it.* Paper presented at the Proceedings of the symposium on eye tracking research and applications.

Hooge, I., Holmqvist, K., & Nyström, M. (2016). The pupil is faster than the corneal reflection (CR): are video based pupil-CR eye trackers suitable for studying detailed dynamics of eye movements? *Vision research, 128*, 6-18.

Nyström, M., Hansen, D. W., Andersson, R., & Hooge, I. (2016). Why have microsaccades become larger? Investigating eye deformations and detection algorithms. *Vision research, 118*, 17-24.

Skovsgaard, H., Agustin, J. S., Johansen, S. A., Hansen, J. P., & Tall, M. (2011). *Evaluation of a remote webcam-based eye tracker.* Paper presented at the Proceedings of the 1st Conference on Novel Gaze-Controlled Applications.

Zhang, B., Stevenson, S. S., Cheng, H., Laron, M., Kumar, G., Tong, J., & Chino, Y. M. (2008). Effects of fixation instability on multifocal VEP (mfVEP) responses in amblyopes. Journal of vision, 8(3), 16-16.

## APPENDIX Modifications of ITU GazeTracker

Listed enhanced capabilities of our version of open-source ITU GazeTracker are provided by following modifications (minor modifications are excluded):

(1) **User Interface** – we added a "Load" button to open pre-recorded video files with calibration and stimulus

display. During replay, gaze position samples are re-obtained and calibration model can be rebuilt. We also added special text fields to type in the physical dimensions of the setup – distance from participant's eyes to computer monitor with stimulus, dimensions of the monitor and others. These measurements are needed to calculate gaze position in °rees of visual angle. Finally, we added a drop list to select which stimulus to use.

(2) **Stimulus** – a new stimulus module, *StimulusExternal.cs* is written to provide programming and displaying various stimuli as dot, images and video stimuli. This module also provide timestamps synchronized with gaze position samples. These timestamps can be provided periodically or for key events only, for example, arriving a dot to a target position or departing from it. For values programmed as random, the actual values are calculated before starting displaying the stimuli. These actual values are logged in an .XML file.

(3) **Camera Control** – a new interface module, *uc480Control.cs* is written to control the eye tracker camera, providing capturing frames in record mode. For this purpose we employ uEye API tools instead of using DirectShow library, as it was done in the original version of ITU GazeTracker. uEye API tools include a software buffer and a sets of commands and system events notifications. Since video streams with high framerate, 120 fps and above generate large files, the modules provides splitting video records into files not larger than 4GB.

(4) **Replaying Video Files** – module *Capture.cs*: we added a new replay mode. We employ DirectShow to provide frame-buffering capability, as well as monitoring. Records that contain more than one video file are replayed seamlessly.

(5) **Lossless Processing** – Module *Tracker.cs*: we added buffering to the original version of the module to prevent frame losses. Initially if the frame could not be immediately processed, it was dropped.

(6) **Saving Gaze Position Samples** – module *Logger.cs*: the original version of ITU GazeTracker was targeted to online eye tracking for pointing and typing purposes and did not provide saving gaze position samples. We added this capability, gaze position samples are saved along with Pupil, P1 position by default and other information.

**Saving Calibration Data** – modules *Calibration.cs* and *Interpolation.cs* are enhanced in order to provide logging of calibration model input data and calculated coefficients.